\documentclass[journal,comsoc]{IEEEtran}
\usepackage[utf8]{inputenc}
\usepackage{color}
\usepackage{acronym}
\acrodef{sn}[SN]{satellite network}
\acrodef{an}[AN]{aerial network}
\acrodef{gn}[GN]{ground network}
\acrodef{rf}[RF]{radio frequency}
\acrodef{ris}[RIS]{reconfigurable intelligent surface}
\acrodef{vlc}[VLC]{visible light communications}
\acrodef{re}[RE]{reflective element}
\acrodef{geo}[GEO]{geostationary orbit}
\acrodef{meo}[MEO]{medium earth orbit}
\acrodef{leo}[LEO]{low earth orbit}
\acrodef{hap}[HAP]{high altitude platform}
\acrodef{lap}[LAP]{low altitude platform}
\acrodef{vn}[VN]{virtual network}
\acrodef{uav}[UAV]{unmanned aerial vehicle}
\acrodef{3d}[3D]{three-dimensional}
\acrodef{fso}[FSO]{free-space optical}
\acrodef{owc}[OWC]{optical Wireless Communications}
\acrodef{bs}[BS]{base station}
\acrodef{ber}[BER]{bit error rate}
\acrodef{thz}[THz]{terahertz}
\acrodef{mmwave}[mmWave]{millimeter wave}
\acrodef{ris}[RIS]{reconfigurable intelligent surface}
\acrodef{mimo}[MIMO]{multiple-input multiple-output}
\acrodef{los}[LoS]{line-of-sight}
\acrodef{nlos}[NLoS]{Non-LoS}
\acrodef{ir}[IR]{infrared}
\acrodef{uv}[UV]{ultraviolet}
\acrodef{ioe}[IoE]{Internet-of-Everything}
\acrodef{occ}[OCC]{optical camera communication}
\acrodef{lifi}[LiFi]{light fidelity}
\acrodef{led}[LED]{light-emitting diode}
\acrodef{ml}[ML]{machine learning}
\acrodef{qos}[QoS]{quality-of-service}
\acrodef{sagin}[SAGIN]{space-air-ground integrated network}
\acrodef{csi}[CSI]{channel state information}
\acrodef{em}[EM]{electromagnetic}
\acrodef{gan}[GAN]{generative adversarial network}
\acrodef{snr}[SNR]{signal-to-noise ratio}
\acrodef{sinr}[SINR]{signal-to-interference-plus-noise ratio}
\acrodef{e2e}[E2E]{end-to-end}
\acrodef{5g}[5G]{fifth generation}
\acrodef{mec}[MEC]{mobile edge computing}
\acrodef{mems}[MEMS]{microelectromechanical systems}
\acrodef{6g}[6G]{sixth generation}
\acrodef{b5g}[B5G]{beyond 5G}
\acrodef{dt}[DT]{Digital Twin}
\acrodef{rl}[RL]{Reinforcement Learning}
\acrodef{tbps}[Tbps]{Tera-bit-per-second}
\acrodef{ct}[CT]{cyber twin}
\acrodef{pt}[PT]{physical twin}
\acrodef{ai}[AI]{artificial intelligence}
\acrodef{ann}[ANN]{artificial neural network}
\acrodef{gnn}[GNN]{graph neural network}
\acrodef{dnn}[DNN]{deep neural network}
\acrodef{rmt}[RMT]{random matrix theory}
 
\usepackage[pdftex]{graphicx}

\title{The Interplay of AI and Digital Twin: Bridging the Gap between Data-Driven and Model-Driven Approaches}  

\begin{document}
\author{Lina~Bariah and M{é}rouane~Debbah
\vspace{-4pt}
}

\maketitle

\begin{abstract} 
The evolution of network virtualization and native artificial intelligence (AI) paradigms have conceptualized the vision of future wireless networks as a comprehensive entity operating in whole over a digital platform, with smart interaction with the physical domain, paving the way for the blooming of the \textit{Digital Twin} (DT) concept. The recent interest in the DT networks is fueled by the emergence of novel wireless technologies and use-cases, that exacerbate the level of complexity to orchestrate the network and to manage its resources. Driven by AI, the key principle of the DT is to create a virtual twin for the physical entities and network dynamics, where the virtual twin will be leveraged to generate synthetic data and offer an on-demand platform for AI model training. Despite the common understanding that AI is the seed for DT, we anticipate that the DT and AI will be enablers for each other, in a way that overcome their limitations and complement each other benefits. In this article, we dig into the fundamentals of DT, where we reveal the role of DT in unifying model-driven and data-driven approaches, and explore the opportunities offered by DT in order to achieve the optimistic vision of 6G networks. We further unfold the essential role of the theoretical underpinnings in unlocking further opportunities by AI, and hence, we unveil their pivotal impact on the realization of reliable, efficient, and low-latency DT. 
\end{abstract}

\section{Introduction}

Over the last couple of decades, the paradigm of virtualization has been evolving from a virtualized local area network (LAN) and private networks to the solidification of network function virtualization (NFV) and network slicing principles. This advancement is driven by the edge computing and cloudification capabilities of current wireless network generations. With the growing demands of wireless networks, in terms of latency, reliability, and energy and spectral efficiency, and the emergence of sophisticated services with heavy distributed computing requirements, it is envisaged that the concept of network virtualization will be scaling up from the node and link levels to the network-wide level, setting the scene for a holistic network virtualization, from the core to the edge. Coupled with the pervasive utilization of \ac{ai} at all network levels, the \ac{dt} paradigm has been recently deemed as a promising tool for network design, optimization, management, and recovery, in which the \ac{dt} can be leveraged to realize the vision of zero-touch 6G networks \cite{khan2022digital}. The key principle of \ac{dt} networks relies on developing an accurate digital replica of a wireless network, taking into account the environmental and physical elements, the network parameters, and the dynamics and interactions happening at the node level. The \ac{dt} paradigm aims at facilitating the optimization and control of wireless networks when implemented at a large-scale. This is motivated by the exacerbated complexity and coordination difficulty of future \ac{6g} networks, which are characterized by the emergence of a swarm of novel applications and technologies, with extreme requirements. Within this context, \ac{dt} is anticipated to offer a digital platform for network configuration and optimization purposes, with the \ac{ai} being the main orchestrator \cite{bariah2022}. Generally speaking, assuming a perfect network and environment virtualization, \ac{dt} can be exploited as a tool for engineered data generation, where data can be collected from common and rare network scenarios, artificially created at the \ac{ct}. This data will be then leveraged by various \ac{ai} algorithms in order to perform models training, and then achieve efficient inference and decision-making process. According to its role, \ac{dt} can be classified into planning, training, and operational twin. The planning twin is utilized at the initial stages to ensure an optimal design of the network assets and components. This means that the \ac{ct} is created before the \ac{pt} in this scenario. Meanwhile, the training twin offers a platform for \ac{ai} models to be trained at the \ac{ct} before implementing them at the \ac{pt}. In this scenario, the computational overhead is moved from the physical environment to the \ac{ct}, and hence, the resources of the physical devices will be saved, and accordingly, the \ac{pt} will be responsible for updating the models in the case of any network variations. Lastly, the operational twin constitutes a network brain, which purpose is to perform data generation, models training based on the generated data at the \ac{ct} and real data collected from the \ac{pt}, and more importantly, perform on-demand decision-making and inference, as well as \ac{ai} models retraining once needed. 

Albeit the prevailing belief that \ac{ai} will be the enabler of the \ac{dt} paradigm \cite{rathore2021role}, it is worthy to divulge whether the contrary is true. Whilst \ac{dt} requires the employment of \ac{ai} algorithms in order to grasp insights from the available data, and hence perform intelligent inference, we envision that the \ac{dt} can potentially contribute to the enhancement of several \ac{ai} algorithms from different perspectives, including the availability of highly reliable data with conditioned distributions, in addition to offering a virtualized digital platform for a reduced complexity at the physical environment. While the former help improving the accuracy of the trained models, the latter can further speed up the training process. Within the same context, recalling that in the current conventional frameworks, due to the lack of high quality data, model-driven approaches are exploited to compensate for the shortage in reliable data and to assist in models training \cite{he2019model}. Although such an approach can temporarily provide a reasonable accuracy, it is unamenable to be scaled up to a large scale, and it lacks the adaptivity to network dynamics. Therefore, we foreseen that the \ac{dt} paradigm will be the link that will provision the synergy of model-driven and data-driven approaches as a unified tool, bringing in their advantages and overcoming their limitations.  

\begin{figure*}
	\centering
	\resizebox{0.8\linewidth}{!}{\includegraphics{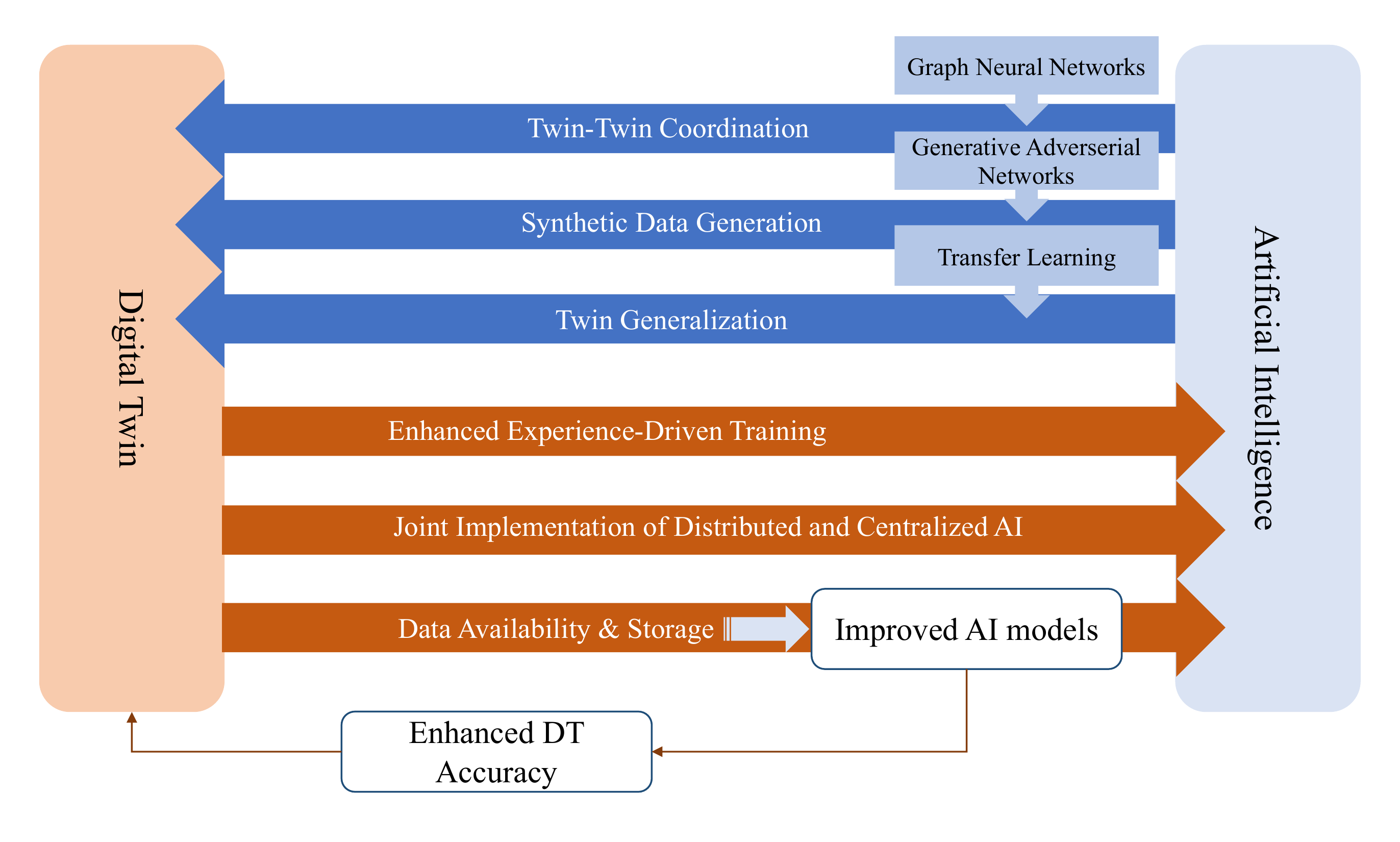}}
	  \caption{AI for DT and DT for AI.}
    \label{fig:Outline}
\end{figure*}

In this article, for the first time in the literature, we explore the interplay of \ac{ai} and \ac{dt}, and delve into the interrelated effect of \ac{ai} and \ac{dt} on each other and how each contributes to the realization of the other paradigm (Fig. \ref{fig:Outline}). Furthermore, we reveal the integral role of model-driven approaches in enabling robust \ac{dt} network, and the intertwined benefits that can be reaped when integrating model-driven and data-driven tools into unified approaches. Also, we shed lights on how the theoretical foundations paves the way for optimized \ac{dt} and for the comprehension of the hidden logic behind most \ac{ai} algorithms. 


\section{AI for Digital Twin}

\subsection{Twin-Twin coordination}

Albeit the bright vision of realizing a holistic representation of the physical environment and network elements over a unified digital platform, the real implementation of \ac{dt} can be achieved through multiple interconnected twins. While latency and complexity will be reduced within each twin, such a distributed \ac{dt} implementation introduces a new level of complexity pertinent to inter-twin coordination, where models trained over the multiple twins should be aligned to ensure accurate global inference, i.e., operations over multiple twins should be synchronized to achieve joint tasks. In this process, reliability and latency should be maintained within the required thresholds. In this context, \ac{ai} algorithms can be leveraged for improved twin-twin coordination. From one perspective, \ac{ai} can be leveraged in order to realize minimized \ac{e2e} latency. In specific, \ac{gnn} constitutes a potential candidate for facilitating inter-twin communication and coordination. This is motivated by the fact that wireless networks modeled through multiple \acp{dt} can be represented as graphs, incorporating the global-context of twins. \ac{gnn} exploits the graph structure of wireless-enabled \acp{dt} in order to capture the nodes dependencies in inter- and intra-twin communications \cite{wang2020graph}. Through the virtual twin, \acp{gnn} can develop comprehensive insights into twins and their interactions by leveraging the dynamic nature of networks as state features, and then aggregate these states to achieve a comprehensive network understanding. Recent results demonstrated the superiority of \ac{gnn} in predicting the network \ac{e2e} delay \cite{almasan2022digital}. This metric can be of importance to accurately measure the delay encountered at the \ac{ct}-\ac{ct} and \ac{ct}-\ac{pt} communications, and compensate for that delay. From another perspective, \acp{gnn} can be exploited for improved data communication and models synchronization among multiple \acp{ct}, where the inter-twin links can be optimized to ensure synchronized twin-twin operations, and ultimately, obtain a homogeneous global \ac{dt}. Additionally, goal-oriented semantic communication has two advantages in twin-twin coordination, namely, i) the proper design of the network goals to achieve the required latency, reliability, and synchronization, and ii) the cooperative operations of multiple twin to fulfill joint network goals, yielding coordinated twins at the virtual realm \cite{wang2023survey}.

\subsection{Synthetic data generation}

\begin{figure*}
	\centering
	\resizebox{0.7\linewidth}{!}{\includegraphics{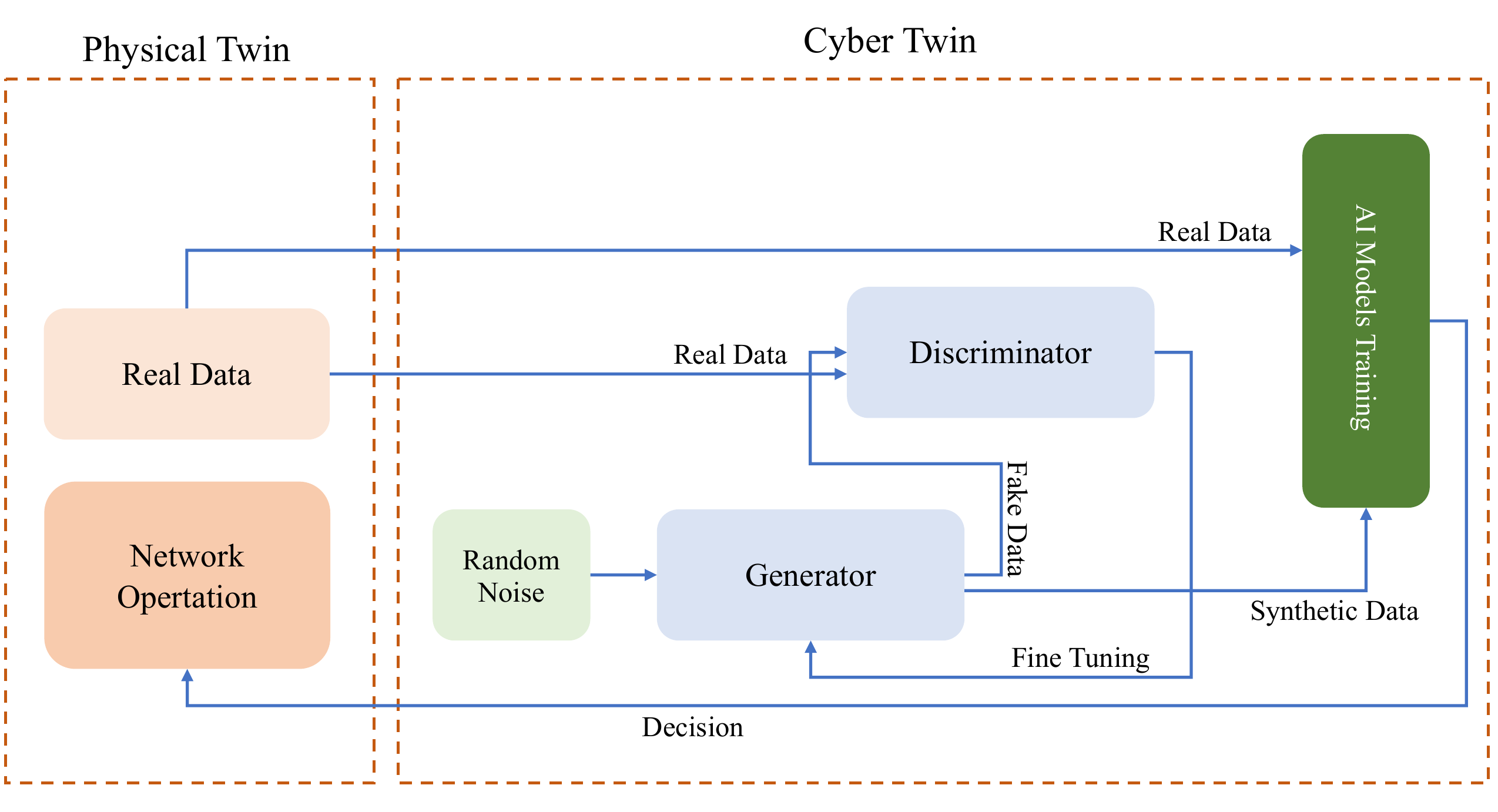}}
	  \caption{GAN-based Digital Twin.}
    \label{fig:GAN}
\end{figure*}

As discussed, one of the attractive benefits of the \ac{dt} is that it constitutes a source for close-to-real data generation, in order to compensate for the weakly measured datasets at edge devices, which are diverse in terms of quality and quantity. Nevertheless, generating all-inclusive dataset that accounts for all nodes status, various environmental events, and network scenarios requires significant time resources, and therefore, is generally performed in a multi-step process. Therefore, \acp{gan} play an important role, by employing a generator and a discriminator in order to generate accurate synthetic data. As demonstrated in Fig. \ref{fig:GAN}, a generator is utilized at the \ac{ct} for the purpose of generating synthetic data. Then, a discriminator is used in order to train the generator to produce higher-quality datasets, i.e., close to the data sensed from the \ac{pt}. Accordingly, the real data acquired from the \ac{pt} is considered as a benchmark to quantify the accuracy of the generated synthetic data from the generator.

Several variants of the \ac{gan} can introduce a different level of data generation options, and hence, offer an engineered data generation campaign, that is aligned with the data generation process at the \ac{dt}. For example, conditional \ac{gan}, which impose particular constraints on the data generated by the \ac{gan} can be leveraged in order to generate datasets that are conditioned by a particular data distribution or different modality \cite{mirza2014conditional}. Furthermore, time-series \ac{gan} can model time-series data \cite{smith2020conditional}, and therefore, can assist with understanding the network dynamics over the \ac{ct}. Note that \acp{gan}, and their variations, have a number of promising applications in wireless networks, e.g., channel estimation/modeling, modulation classification and recognition, and spectrum management. Accordingly, \acp{gan} represent an essential part in DT-enabled wireless networks, where synthetic data generated by \acp{gan} are expected to substantially enhance the performance of corresponding wireless networks. Within this regard, \acp{gan} at the cyber twin are anticipated to generate data with different modalities, according to the need of the considered scenario, and hence, images, RF data, etc, are potential outputs of \acp{gan} at the \ac{dt}. It is worthy to note that data generated using \acp{gan} can be used to validate models trained over the \ac{dt} data, and vice versa.

\subsection{Twin generalization}

As discussed earlier, due to the increased complexity associated with future wireless networks, distributed \acp{dt} might be the solution for enhanced reliability and reduced latency in the \ac{dt} paradigm. While on-demand data sensing is one of the operational \ac{dt} pillars, it might be challenging and time-consuming to perform models retraining as a response for any environmental or network variations. This is particularly pronounced in sudden and fast variations that are generally expected in vehicular networks. Accordingly, transfer learning can be exploited to propagate the network updates over multiple twins, yielding reduced latency and overhead. Transfer learning can be applied to multiple \acp{dt} scenario, where retrained models, due to network variations, over a particular distributed twin, can be communicated with other twins with the aim to reduce the time and computing resources needed for the new models training \cite{pan2009survey}. This will further enable a more generalized distributed \ac{dt} frameworks, where the knowledge in a single twin, can be applicable to other twins. While the advantages of transfer learning are particularly appealing in models retraining, it can be exploited at the initial training process of multiple twins, especially when multiple twins are experiencing noticeable variations in the measured and generated data sizes and quality. In this setup, models trained at twins with sufficiently available datasets can be communicated with other twins with the purpose of speeding up the training process at target twins. In specific, a \ac{ct} with limited dataset (due to unreliable physical-cyber communication) can benefit from a well-trained model from another \ac{ct}. Assuming related tasks between two \acp{ct}, the knowledge obtained from the first task training, which is based on a large and high-quality dataset, can be generalized to the second task at the other twin. 

Note that transfer learning can be integrated with \ac{rl} for improved \ac{ct} design. \ac{rl} can be exploited for improved knowledge transfer between multiple \acp{ct}, where the transferred model from a well-established \ac{ct} can be exploited at policy-agnostic agents in another \ac{ct}, which experiences low-quality dataset, hence, rewards and actions will rely on the transferred model, and thereby, the transferred models' weights can be fine-tuned to optimally fit the target twin. On the other hand, transfer learning can be used for policy transfer among multiple cyber twins, for fast convergence and agents training at the target \ac{ct}.

\section{Digital Twin for AI}

While the successful operation of \ac{dt} heavily relies on various \ac{ai} algorithms, it is worthy to ask whether the opposite is true. Will the development of a holistic \ac{dt} be the seed for more innovative \ac{ai} architectures, that will serve the interests of future wireless networks? Furthermore, can \ac{dt} be leveraged to enhance existing approaches, both model-driven and data-driven? If so, how \ac{dt} can be efficiently implemented in order to enable it to fully grasp the benefits of both approaches? Recalling that future wireless networks are characterized by their high level of complexity, it is indisputable that at some point current schemes will fail to deliver the needed performance and accuracy. This is due to the fact that existing network optimization, configuration, and design are either developed from a theoretical point of view, or built based on data collected from the network. While the former is efficient in initial network planning, it lacks the scalability and adaptivity offered by the latter. On the other hand, data-based models are insufficient to provide full-scale representation of wireless networks. In the following, we explore the opportunities offered by the \ac{dt} in order to provide a unified platform for model-based and data-based schemes.   

\subsection{Experience-driven learning} 

As \ac{rl} was initially developed as a step toward realizing autonomous systems, it constitutes a natural choice for \ac{dt} applications. 
The merit of \ac{rl} is manifested in training \acp{dt}, where a training \ac{dt} can be employed for risk-controlled agents training, i.e., \ac{rl} agents can freely interact with the CT, experiencing a wide range of common and rare scenarios. While maintaining the real environment unharmed is considered a big advantage, \ac{rl} agents can further benefit from the \ac{dt} for fast training purposes, where devices with super-computing capabilities can ensure accurately trained agents at the \ac{ct} within a short period of time. These benefits are further demonstrated through a case study by Ericsson, in which they developed a \ac{dt} framework to minimize the transmission power, while maintaining a particular \ac{qos} requirement and a monitored level of \ac{rf} radiations \cite{Erc-RL}. Allowing the \ac{rl} agents to learn through direct interaction with the physical environment is considerably risky, particularly in areas with strict regulations on transmission power, and therefore, the \ac{dt} offers a safe, yet efficient, virtual agents training. The optimum goal of such scenarios is to design multiple agents that are well-trained in a way that enables them to perform efficiently without further interactions with the physical environment, or to require few interactions with the real environment. 
\begin{figure*}
	\centering
	\resizebox{0.85\linewidth}{!}{\includegraphics{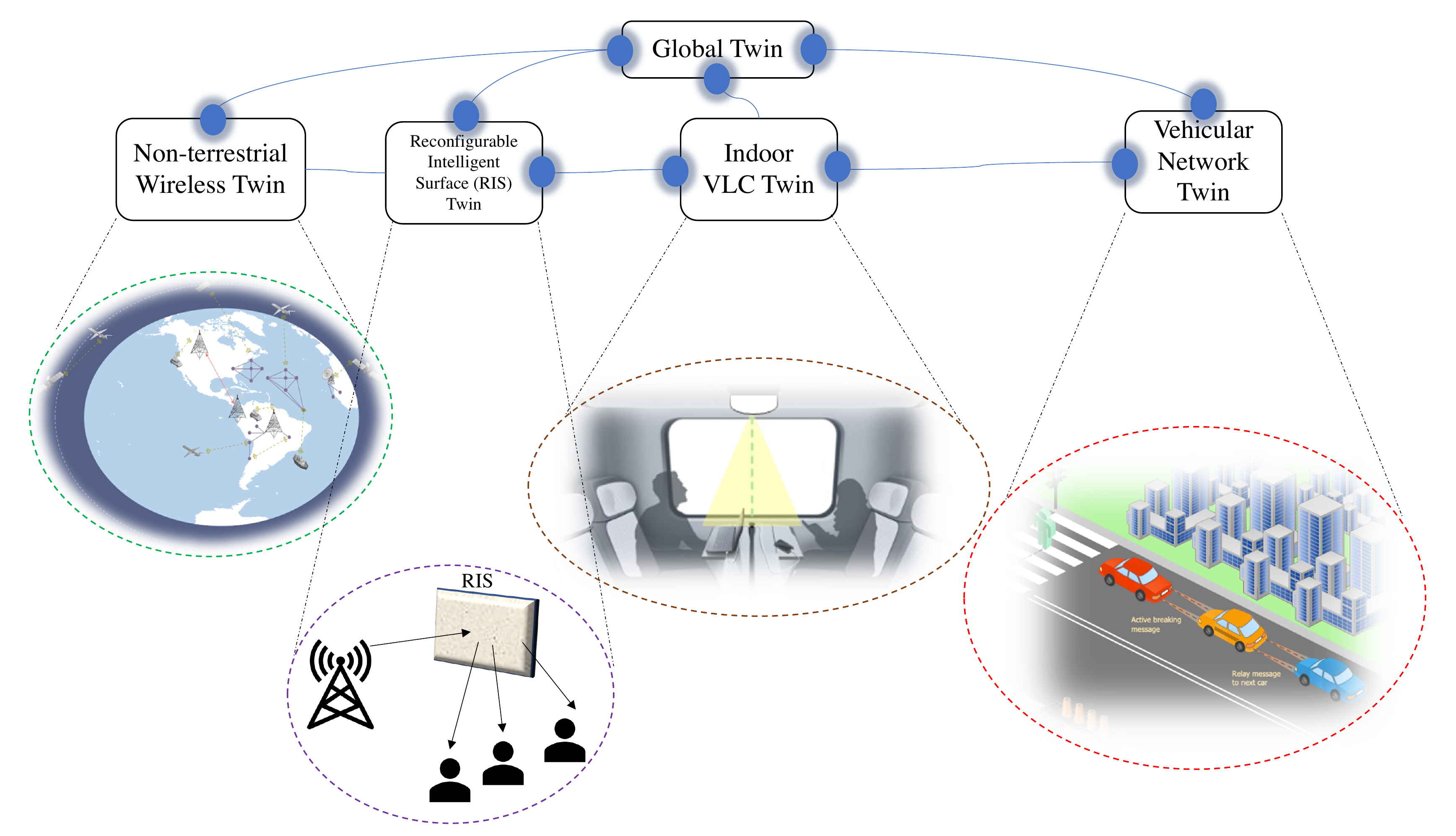}}
	  \caption{GNN-based Twin-Twin Coordination.}
    \label{fig:GNN}
\end{figure*}

\subsection{Data availability and Storage} 
\label{DA}
Although the recent the advancement in sensing services has facilitated data measurements campaigns, for improved \ac{ai} models training process, the envisioned native \ac{ai} networks will necessitate on-demand node-level data collection. This is primarily aimed in order to enable pervasive intelligence and accurate inference regardless of the network status. Such a bright vision can only be achieved if data collected represents all possible network scenarios, taking into account the physical environment status. Therefore, the \ac{dt} can be a game-changer in this situation, where rarely-experienced network scenarios can be artificially engineered at the \ac{ct} in order to study the network behaviour under such circumstances, and hence, perform comprehensive data collection process, that is capable of representing all nodes activities under a wide-range of network scenarios. While this data is artificially generated, it is considered close-to-real data, given that it will be generated under realistic virtual environments, that accurately imitate the dynamics of real networks. Within the same context, recalling that future wireless networks are characterized by their high level of heterogeneity, it is highly probable that local datasets at edge devices are non-identically distributed and differ in quantity and quality. This in consequence will result in models uncertainties, and hence, severely impact the network performance. \Ac{dt} in this context ensures that models are trained over highly-reliable data in terms of quality and quantity. Note that, in order to guarantee a general-enough and accurate models when \ac{ml} algorithms are executed in a supervised fashion, data used for testing should be different than the data used for training, however, both should be drawn from similar distributions \cite{zappone2019wireless}. This further corroborates the role of \ac{dt} in empowering highly reliable and efficient \ac{ml} algorithms, where not only large datasets can be generated, but their distributions can be controlled to ensure the required \ac{qos}. 

\subsection{Virtual implementation of AI: Unifying distributed and centralized approaches}
Several research activities were initiated to explore the merit of \ac{ai} when implemented in a distributed fashion. These activities were fueled by the increased communication overhead and latency, and compromised privacy, which are generally experienced in centralized methods. Although the research on distributed AI has picked up the pace in the recent years, it is still uncertain whether edge nodes are qualified for delivering the required \ac{qos}, particularly with the emergence of native \ac{ai} concept, where each network node is anticipated to perform sensing, training, and inference at some level. In this regard, the \ac{dt} offers a robust platform to calibrate the virtue of distributed and centralized algorithms, and to overcome their limitations. On the one hand, the overhead resulted from datasets exchange between the edge devices and the centralized aggregator will be alleviated from edge devices, users' privacy will be maintained, and high latency encountered in centralized algorithms will be reduced. On the other hand, low training accuracy in distributed schemes, due to the local datasets limitations, can be significantly improved through the implementation of the \ac{dt} paradigm, where all-inclusive datasets are available for enhanced models training. Distributed models trained at the \ac{ct} might require necessary updates once implemented at the physical environment, which can be done at the edge devices at the \ac{pt}, and hence, the role of edge devices will be limited to updating the local models according to the new circumstances at the physical environment. Meanwhile, in the event of operational interactive twin, models updates can be performed at the \ac{dt} as well, yielding a latency-accuracy-energy trade-off problem. 
\begin{figure*}
	\centering
	\resizebox{0.7\linewidth}{!}{\includegraphics{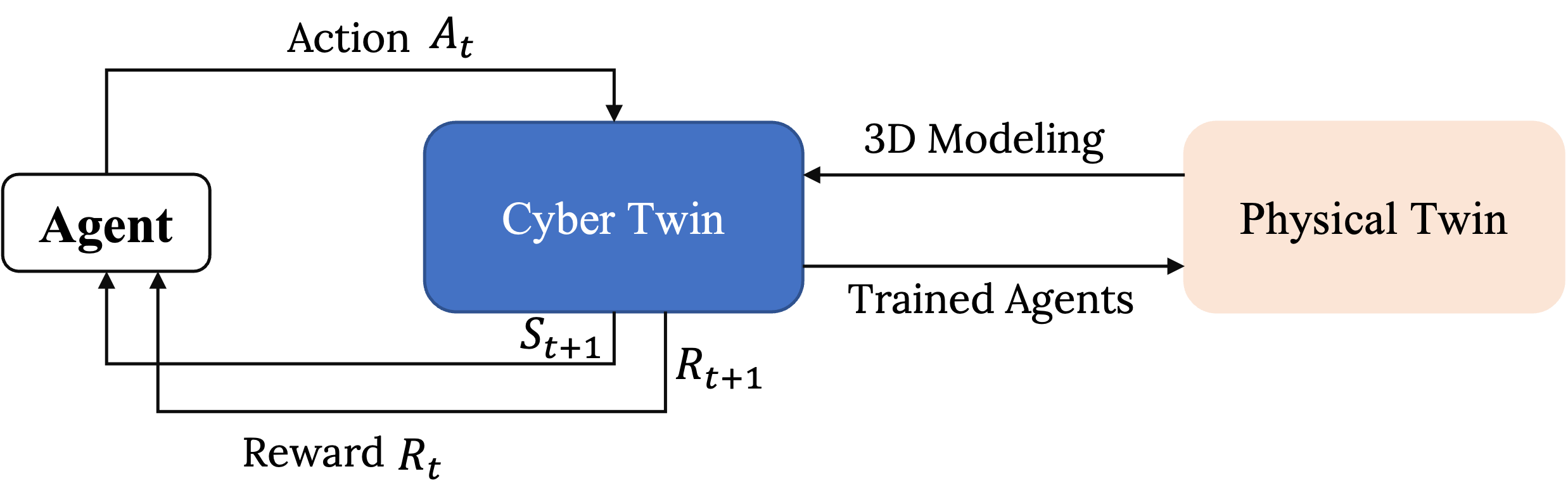}}
	  \caption{DT-enabled reinforcement learning.\vspace{-6pt}}
    \label{fig:RL}
\end{figure*}
\vspace{-4pt}
\section{Model-based Digital Twin}

With the advent of \ac{ai} and its promising advantages, it is now unusual to envision wireless networks without an \ac{ai} element, where the employment of \ac{ai} algorithms has become a trivial solution for any problem in wireless networks. While we fully agree that \ac{ai} is an indispensable tool in future wireless generations, we cannot relegate the integral role of model-driven approaches and the advantages that they can bring to the network design and optimization process. Although the common understanding of the \ac{dt} paradigm is confined by creating a virtual replica of an existing physical domain, it is not a necessary for the \ac{pt} to be available in order for the \ac{ct} to be created \cite{madni2019leveraging}. In the latter scenario, model-based approaches represent the key for the preliminary design of network assets and to support the decision-making process at the initial stages. The focus of this type of \acp{dt} is to mitigate technical risks, through exploring the network behaviour at the \ac{ct} under what-if analysis mode.

On the other hand, mathematical models, not only facilitate the understanding of the logic behind \ac{ai} algorithms, but also provides a resilient abstraction for the \ac{dt} assets and dynamics. Therefore, it is insufficient to completely focus on \ac{ai}-driven approaches to realize efficient \ac{dt} network. Rather, efforts should be devoted to developing solid mathematical underpinnings in order to be able to identify the theoretical limitations of \ac{dt}, put the foundations for a comprehensive mathematical interpretation of \ac{ai} algorithms, and accordingly unleash the full potential of \ac{dt} \cite{MB}. Among several mathematical theories, we believe that the optimization theory, random matrix theory, graph theory, optimal transport theory, stochastic geometry, and game theory are essential tools that are required for an efficient construction and operation of digital twins. Such tools constitutes the base for i) modeling the randomness of physical environments and electromagnetic signals, taken into consideration the large amount of data to be sensed and communicated between the cyber and physical twins, ii) synchronizing and optimizing the operations among multiple cyber twins, particularly in a massive twinning scenario, to ensure a harmonized global twin, iii) balancing and coordinating the association and decoupling of multiple digital twins, and iv) robust 3D modeling of wireless networks, taking into consideration the spatial components. 

\subsection{The fusion of model-driven and data-driven approaches over the twin}

The role of the mathematical frameworks in the design and optimization of wireless networks cannot be completely ignored, and a full reliance on \ac{ml} tools will leave a noticeable gap. It is worthy to note that mathematical frameworks constitute solid pillars that pave the way for \ac{ml} algorithms to achieve enhanced performance and scalability \cite{chafii2022ten}. Therefore, despite the advanced progression in \ac{ai} models, it is an imperative fact that the absence of model-driven approaches in the design and optimization process represents a bottleneck in future wireless networks. Model-driven approaches faces two major limitations: i) In some complex network scenarios, e.g., ultra-high dense heterogeneous networks, the resulted expressions representing the system are mathematically intractable, and the associated optimization problems do not lend themselves into closed-form optimum or sub-optimum solutions, and ii) The reduced accuracy when the available mathematical framework cannot be readily scaled up to represent another network scenario. Accordingly, the \ac{dt} offers a way out through integrating the benefits of model-driven and data-driven approaches in a unified process. In particular, the intractability issue of some model-driven approaches can be solved through employing \acp{ann}, which is a sub-field of \ac{ml}, in a way that allows \acp{ann} to map the network parameters into the corresponding network performance, in a numerical fashion, until the convergence to the optimum solution. Such exhaustive approach is heavy and might not be tolerated by nodes with limited resources. Thanks for the \ac{dt} paradigm, such a complexity can be alleviated from the physical nodes, yielding robust models training, that enjoys a solid mathematical foundation, with reduced overhead from the \ac{pt}. Note that the role of \ac{dt} in this scenario is to offer a platform for low-complex integration of theoretical models with \ac{ml} frameworks. On the other hand, to tackle the inaccuracy limitations associated with network scalability in model-driven approaches, conventional approaches rely on initially training \ac{ml} algorithms using the available mathematical models. Then, through synthetic data generation, trained models will be refined. In this scenario, the \ac{dt} ensures the availability of sufficient, accurate, and close-to-real datasets for improved models refinement, and therefore, enhanced accuracy.  

\section{Challenges and Future Directions}


\subsection{Exact digital replica}
It is not hard to tell that the merit of the \ac{dt} as enabler for \ac{ai} is confined by the accuracy and the synchronization between the \ac{ct} and \ac{pt}. As discussed earlier, the benefits of the \ac{dt} are revolving around acquiring an exact replica of the physical objects, and the instantaneous dynamics of the environment and network parameters, over a long period of time. This is due to the need of exploiting the \ac{dt} as a digital environment for data generation and models training, and therefore, its accuracy is highly dependent on how accurate and realistic the \ac{ct} compared to the \ac{pt}. This opens the doors to further investigate the limitations of a holistic network virtualization, and to identify potential key solutions.   



\subsection{Fast AI algorithms}

While the availability of large datasets at the \ac{ct} for the purpose of \ac{ai} models training is appealing, it is debatable whether current \ac{ai} architectures will perform in a reliable and timely manner. Recalling that ultra-low latency is one of the \ac{dt} verticals, it is essential to ensure that employed \ac{ai} algorithms will be able deliver the required accuracy at a time-frame that is tolerable by the \ac{dt} paradigm. The same applies to models updates at the \ac{ct}. Hence, is there a need for designing new \ac{ai} algorithms for the implementation of efficient \ac{dt}? If not, to what extent current \ac{ai} algorithms will be able to meet the latency requirements of the \ac{dt}? This will further require the exploration of data management, cleaning, and rectifying methods, in order to enable efficient execution of \ac{ai} algorithms. 

\subsection{Unlocking the theory behind AI algorithms}

Although the benefits of \ac{ai} algorithms are particularly pronounced in highly dynamic, large-scale twins, such algorithms are generally treated as black boxes, where the tangible insights are obtained from the input and output data, while the complex operations in between, which are generally characterized by their extremely high complexity, remain difficult to understand, modify, and improve. In this regard, the proper understanding of the hidden operations in \acp{dnn} paves the way for improved architectures that better serve the needs of \ac{dt}. As a promising approach, \ac{rmt} has manifested itself as a versatile, yet solid, approach for analyzing \acp{dnn}. Note that, coupled with the large non-identically distributed data generated from the \ac{ct}, through randomly initializing the parameters of a neural network, complex \acp{dnn} can be represented as a large matrix of random variables. Accordingly, \ac{rmt} can be leveraged to optimize the initial weights of \acp{dnn}, particularly when implemented on large-scale. In specific, \ac{rmt} can be efficiently exploited to design novel activation functions, that will play a role in speeding up the training process, and therefore, realize low-latency \ac{dt} \cite{chafii2022ten}. 

\subsection{Digital Twin-empowered emergent intelligence}

The proliferation of machine-to-machine communications have stimulated the recent interest in contextual-based decision-making, i.e., the so-called emergent intelligence, where multiple agents collaborate to perform a particular task through intensively interacting with each other and with the environment. 
While wireless networks offer an environment of a large number of communicating agents, and hence, enable the agents to develop a more advanced language that is capable of conveying a variety of abstract ideas, such scalability feature will result in either an increased latency or reduced reliability due to the substantially increased communication traffic. Within this context, \ac{dt} represents an optimum candidate to enable reliable agents interaction and to support the deployment of emergent communication in future wireless networks. In particular, digital replicas of the communicating agents can be implemented at the \ac{ct}, yielding improved agents information exchange, reduced latency, enhanced links reliability, reduced signaling overhead, while ensuring accurate inference \cite{khan2022digital}. As \ac{dt}-empowered emergent intelligence is still in its infancy, it is essential to understand how \ac{dt} can leverage the semantic aspect of communicated messages to develop a common level of understanding between communicating digital agents, that enables them to perform assigned tasks successfully, and to reflect this learning and inference process to the physical agents.

\section{Conclusion}

Despite the trending concept that \acp{dt} will be completely controlled by \ac{ai}, in this article we elucidated the interplay of \ac{dt} and \ac{ai} as being enablers and enabled by each other. Furthermore, we revealed the uncommon opinion that model-driven approaches have an essential role in the realization of efficient and accurate \ac{dt}. In particular, we offered a forward-looking vision on how model-driven tools will complement data-driven approaches and assist with overcoming their limitations. We further emphasized on the indispensable role of the mathematical frameworks on understanding, improving, and designing \acp{dt}, as well as the importance of leveraging theoretical frameworks for enhanced \ac{ai} architectures. 
It is worthy to highlight that, although several AI algorithms can be devoted to realize a high-quality \ac{dt}, the full potential of \ac{dt} can be achieved through the amalgamation of multiple \ac{ai} algorithms over the twin, where further advantages, pertinent to latency, complexity, and reliability can be reaped by unifying the multiple algorithms in the \ac{dt}.

\bibliographystyle{IEEEtran}
\bibliography{Ref}

\section*{Biographies}

\textbf{Lina Bariah} (lina.bariah@ieee.org) is a Senior Researcher at the Technology Innovation Institute in Abu Dhabi. She is an IEEE Senior Member. She serves as an Associate Editor for the IEEE Communication Letters, and the IEEE Open Journal of the Communications Society.  

\textbf{M{\'e}rouane Debbah} (Merouane.Debbah@tii.ae) is the Chief Researcher at the Technology Innovation Institute in Abu Dhabi. He is an IEEE Fellow, a WWRF Fellow, a Eurasip Fellow, an AAIA Fellow, an Institut Louis Bachelier Fellow, and a Membre émérite SEE. He has received more than 20 best paper awards.

\end{document}